\documentclass[letterpaper, 10 pt, conference]{ieeeconf}

\IEEEoverridecommandlockouts                              

\usepackage{dsfont}
\usepackage{lipsum}
\usepackage{amsmath}
\usepackage{graphicx}                       
\usepackage{stfloats}
\usepackage[tight,footnotesize]{subfigure}  

\usepackage{amssymb,amsmath}
\usepackage{textcomp}
\usepackage{mdwmath}
\usepackage{mdwtab}
\usepackage{eqparbox}
\usepackage{rotating}                       
\usepackage{array}                          
\usepackage{listings}                       
\usepackage[ruled,vlined,linesnumbered]{algorithm2e}
\usepackage{listings}
\usepackage[noend]{algpseudocode}
\usepackage{CJK}
\usepackage{indentfirst}
\usepackage{amsmath}
\usepackage[colorlinks,
            linkcolor=blue,
            anchorcolor=blue,
            citecolor=blue,
            bookmarks=false
            ]{hyperref}
\usepackage{cite}

\usepackage{graphicx}
\usepackage{float}
\usepackage{subfigure}
\usepackage{amsmath}
\usepackage{algpseudocode}
\usepackage{stfloats}
\usepackage{color}

\usepackage{multirow} 
\usepackage{xcolor}
\usepackage[ruled,vlined]{algorithm2e}
\usepackage{amssymb}
\usepackage{booktabs}
\title{\LARGE \bf A Data-efficient Framework for Robotics Large-scale LiDAR Scene Parsing}
\author{Kangcheng Liu,~\IEEEmembership{Member,~IEEE}, and Ben M. Chen,~\IEEEmembership{Fellow,~IEEE}%
\thanks{K. Liu and B. M. Chen are with the Department of Mechanical and Automation Engineering, The Chinese University of Hong Kong, Shatin, N.T.,  Hong Kong 999077, China.
(email: {\tt\small kcliu@mae.cuhk.edu.hk, bmchen@cuhk.edu.hk)}}
\thanks{Z. Gao is with the School of Remote Sensing and Information Engineering, Wuhan University, Hubei 430070, China.
(email: {\tt\small gaozhinus@gmail.com)}}
}   

\begin{document}

\maketitle
\thispagestyle{empty}
\pagestyle{empty}

\vspace{-0.1mm}
\begin{abstract}
Existing state-of-the-art 3D point clouds understanding methods only perform well in a fully supervised manner. To the best of our knowledge, there exists no unified framework which simultaneously solves the downstream high-level understanding tasks, especially when labels are extremely limited. This work presents a general and simple framework to tackle point clouds understanding when labels are limited. We propose a novel unsupervised region expansion based clustering method for generating clusters. More importantly, we innovatively propose to learn to merge the over-divided clusters based on the local low-level geometric property similarities and the learned high-level feature similarities supervised by weak labels. Hence, the true weak labels guide pseudo labels merging taking both geometric and semantic feature correlations into consideration. Finally, the self-supervised reconstruction and data augmentation optimization modules are proposed to guide the propagation of labels among semantically similar points within a scene. Experimental Results demonstrate that our framework has
the best performance among the three most important weakly
supervised point clouds understanding tasks including semantic
segmentation, instance segmentation, and object detection even
when limited points are labeled, under the data-efficient settings for the large-scale 3D semantic scene parsing. The developed techniques have postentials to be applied to downstream tasks for better representations in robotic manipulation and robotic autonomous navigation. Codes and models are publicly available at: https://github.com/KangchengLiu. 

\end{abstract}

\section{Introduction}
Previous point clouds understanding methods rely on heavy annotations, and understanding of large-scale 3D scenes requires a large amount of high-quality labels, which are commonly unavailable or labour-intensive to achieve in real situations. For example, labeling a scene in S3DIS or ScanNet requires hundreds of annotators, and takes approximately half an hour per scene for thousands of scenes. For large-scale common indoor/outdoor scenes in robotics interaction and autonomous driving, it becomes unrealistic. Therefore, weakly supervised learning (WSL) based 3D point clouds understanding is highly in demand. Motivated by the success of WSL in images, many works start to tackle weakly supervised understanding with fewer labels, but great challenges remain. In general, the previous methods suffers from a lot of limitations, including heavy annotation cost for semantic labeling of images projected from point clouds as well as information loss, complicated pre-processing and pre-training, customized labeling strategy for sub-clouds, lack of relationship mining both in low-level geometry and high-level semantics.  Therefore, there is a lot of room to explore how to fully unleash the capacity of WSL to make full use of weak labels and mining semantic /geometric correlations among weakly labeled and unlabeled regions.  

\begin{figure}[t]
\setlength{\abovecaptionskip}{-0cm}
\setlength{\belowcaptionskip}{-0cm}
\centering
\includegraphics[scale=0.46]{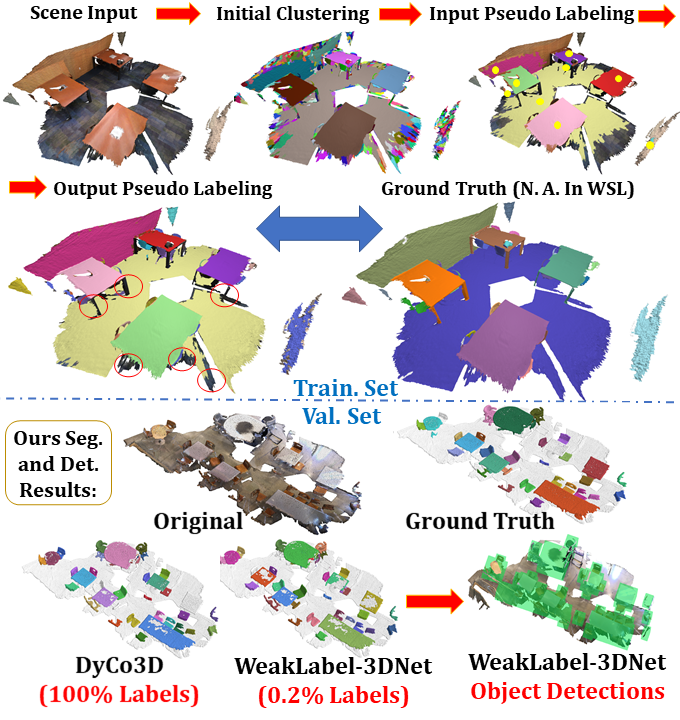}
\caption{Above the dash line shows our learning based pseudo label generation by region expansion. The ground truth is not available in weakly supervised learning (WSL). Below shows our instance segmentation compared with current SOTA DyCo3D and object detection results respectively.}
\label{Intro}
\vspace{-0.288cm}
\end{figure}

Motivated by challenges in data-efficient 3D scene understanding, we study how to take advantage of limited labels to realize multi-tasks point clouds WSL involving 3D semantic segmentation, instance segmentation, and object detection. As illustrated in Fig.~\ref{Intro}, we propose an unsupervised method for generating initial clusters by region expansion based on local normal and curvature, which are the two most prominent geometric characteristics of 3D objects. Next, we put forward a learning-based cluster-level similarity prediction network to predict the similarities among clusters in the latent space. Then the region merging is applied again to aggregate similar clusters guided by both low-level geometric and learned high-level semantic relationships to output pseudo labels. We design self-supervised learning schemes to optimize the network with data augmentation and reconstruction losses to propagate the weak labels to semantically similar regions. Unsupervised learning of instance segmentation can be achieved based on proposed modules to provide supervisions for object detection.
Our proposed framework attains comparable segmentation performance with existing fully supervised state of the arts (SOTAs) and significantly outperforms current weakly supervised SOTAs. To the best of our knowledge, our work is the first unified framework to tackle the weakly supervised multi-tasks 3D point clouds understanding.




Here we summarize several prominent contributions of our work:
 \begin{enumerate}

\item We propose an unsupervised method for generating clusters based on point clouds local geometry, and put forward a learning-based cluster-level similarity prediction network to aggregate similar clusters guided by similarities in both low-level geometry and high-level semantics.

\item We propose self-supervised 
data augmentation and reconstruction optimization functions to make the utmost of weak labels by propagating them to similar points in latent space. 

\item We innovatively propose to provide supervision for object detection by unsupervised learning of instance segmentation to generate bounding boxes pseudo labels.
Regression losses are put forward to take advantage of pseudo labels.

\item Top-ranking performance has been achieved by our framework with extensive experiments on publicly available ScanNet benchmarks and lots of other indoor/outdoor benchmarks including S3DIS, KITTI, Waymo, and Sensaturban with diverse experimental circumstances. Our comprehensive results have provided baselines for future research in 3D WSL.

 \end{enumerate}
\vspace{-0.01cm}
  \section{Related Work}
  \subsection{Learning Based Point Clouds Understanding Approaches}
Deep network based approaches are widely adopted for point clouds understanding. The fully supervised approaches can be roughly categorized into voxel-based, projection-based and point-based methods, Many recent work propose to pre-train networks on source datasets with an auxiliary task such as registration or completion, and then transfer and finetune network weights with the contrastive learning for the aimed 3D understanding tasks to boost performance. However, all above methods requires accessibility to fully labeled ground truth.   

\vspace{-0.1cm}
\subsection{Weakly Supervised methods for Point Clouds Understanding}
Transforming point clouds to images is a great choice for obtaining semantic map, but image-level labels are required for training. Sub-cloud annotations requires the extra labour to separate sub-clouds and to label points within the sub-clouds.  Directly extending current art methods with weak labels for training will result in a great decline in performance if label percentage drops to a certain value which is less than 1\textperthousand\.. Self-training techniques have been utilized to design a two-stage training scheme to produce pseudo labels from weak labels, but it is only tested for the semantic segmentation task with limited performance. Xu et al. adopts semi-supervised training strategies combining training with coarse-grained information and with partial points using on tenth labels, but their test datasets are limited and it is tough to uniformly choose points to label. The unsupervised pre-training shows great capacity in unleashing the potentials of weak labels to serve for complicated tasks, such as instance segmentation. But merely utilize pre-training can not make full utilization of the weak labels, which results in dissatisfactory performance.  
 
 \vspace{-0.2cm}
\subsection{3D Semantic/Instance Segmentation and Object Detection}
Recent studies have produced many elaborately designed networks for 3D semantic/instance segmentation and object detection. However, they all rely on full supervisions. In addition, many frameworks focus only on a single task or two similar tasks, and the relationships mining between those interconnected or complementary tasks, such as correlations between 3D instance segmentation and object detection, are rarely explored. 
\vspace{-0.1cm}
 \section{Proposed Methodology}
We propose a general framework to tackle weakly supervised 3D understanding. A novel unsupervised region expansion clustering to obtain initial pseudo labeled clusters is proposed in Subsection A. Our network framework and innovatively designed modules to merge over-divided clusters to provide pseudo labels are illustrated in Subsection B and C for segmentation and detection respectively. Also, the LiDAR-based approaches are of significance to many industrial applications such as UAV, UGV as well as service robotic navigation as well as inspections~\cite{liu2022robustmm, liu2022industrialTIE, liu2022semi} and robotic enhanced large-scale localization in the diverse complex environments~\cite{liu2022light, liu2022weaklabel3d, liu2022robustcyb, liu2022robust, liu2022integratedtrack, liu2023dlc, liu2022enhanced, liu2022enhancedarxiv, liu2022lightarxiv}, and large-scale robotic semantic scene parsing~\cite{liu2021fg, liu2022fg, liu2020fg}, as well as robotic control as well as robotic manipulation applications~\cite{liu2017avoiding, liu2023lidar, liu2022integrateduav, liu2022integratednoise, liu2022datasetsicca, liu2023learning}, etc.
\subsection{The (Learnable) Similar Region Expansion for Points Clustering}
\vspace{-0.1cm}
\begin{algorithm}[bp]
\footnotesize
\label{alg_clustering}
\setlength{\abovecaptionskip}{-0.35cm}
\setlength{\belowcaptionskip}{-0.35cm}
     \caption{The (Learnable) Similar Region Expansion for Points Clustering} 
      \KwIn{The \textbf{input} raw point set $\textbf{P}_i=\{p_i\}, i=1, 2, ..., N_{i}$. $p_i=(x_i, y_i, z_i)$.} 
      \KwOut{The \textbf{output} pseudo label matrix \textbf{\textit{L}} for different clusters.}
  Initialize  $P_{seed}=p_{\{seed, i\}}, i=1, 2, ..., N^{seed}$\; Initialize the pseudo label matrix \textbf{\textit{L}} as a zero matrix.
  
  \While{not converged}
{ Select $K$ nearest neighbour points $p_i$ around    the seed points $p_{seed}$ for comparisons based    on fast Octree-based $K$ Nearest Neighbor Search\;
        
        \For{the seed points $p_i$ selected} 
        {
            \If{\textbf{Condition 1}}
            {Assign $p_i$ the same class label as $p_{seed}$\;
            
                \If{\textbf{Condition 2}}
                  {Regard the point $p_i$ as new seed points\;}
            }
             \Else{Assign $p_i$ with a new class label. Regard  $p_i$ as new seed points;} 
             $i \leftarrow i+1$\;
             
             Update the pseudo label matrix \textbf{\textit{L}}.}
        } 
\Return The class label matrix $\textit{\textbf{L}}$ of $\textbf{P}_{i}$ with different clusters.
\vspace{-0.05cm}
\end{algorithm}

\begin{figure}[bp]
\setlength{\abovecaptionskip}{-0.15cm}
\setlength{\belowcaptionskip}{-0.15cm}
\centering
\includegraphics[scale=0.180]{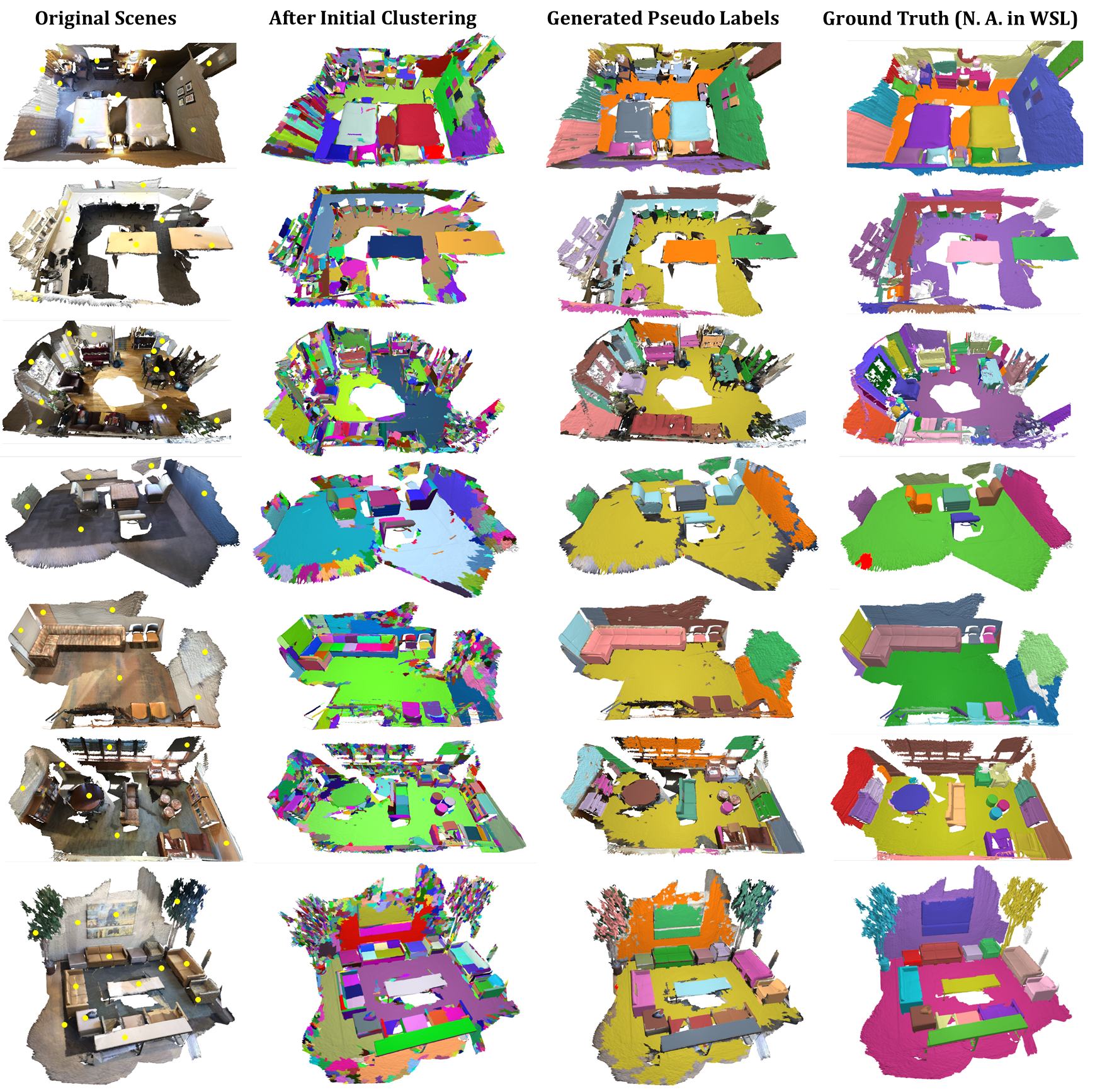}
\caption{Region expansion based labeling results and comparisons with ground truth, which is not available in weakly supervised learning. Truly labelled points are indicated by yellow in the first column of original scene.}
\label{Fig_Region_Expansion}
\label{Clust}
\end{figure}
We provide the initial clustering based on the local geometric properties including normal and curvature of points. First, the Octree based K nearest neighbour (KNN) search is conducted for the acceleration of nearest neighbour query process. Improved based on efficient PFH-based segmentatation approaches, we proposed the following three criterion for a faster KNN query. \textbf{1}. If an octant is not overlapping with the query ball, we should skip it. \textbf{2}. If the query ball is inside an octant, we stop searching. \textbf{3}. If the query ball contains the octant, we just compare the query with all points, so going into children of that octant is not required. We greatly improve the query speed by 18.2 times, and it also largely speeds up the normal and curvature calculations. 


Then, points are ranked according to curvatures. The truly labeled points and points that have top 2\textperthousand\,  minimum curvature among all points are regarded as seed points. The Octree based KNN search of seed points is also adopted in each iteration for acceleration. Denote the input raw point set as $\textbf{P}_i=\{p_i\}, i=1, 2, ..., N_{i}$, $p_i=(x_i, y_i, z_i)$, where $N_{i}$ is the number of input points. Representing the normal vectors of $p_i$ and $p_{seed}$ as $\textbf{n}_i$ and $\textbf{n}_{seed}$, and the curvature of $p_i$ and $p_{seed}$ as $r_i$ and $r_{seed}$\;. We calculate the angle $\Delta \phi$ between $\textbf{n}_i$ and $\textbf{n}_{seed}$, and the difference in curvatures $\Delta r$ between $r_i$ and $r_{seed}$\;, as shown in Algorithm \ref{alg_clustering}. We design \textbf{\textit{Condition 1}} and \textbf{\textit{Condition 2}} for normal and curvature respectively depending on the properties of datasets. We set the \textbf{\textit{Condition 1}} as: $\Delta \phi \leq \gamma$, and \textbf{\textit{Condition 2}} as: $ \Delta r \leq \sigma$. In our experiments, take indoor case for example, we set $\gamma=2.2^{\circ}$ and $\Delta r= 0.35 m^{-1}$ based on geometric properties and dimensions of ScanNet and S3DIS. We set the point as seed point only when \textbf{\textit{Condition 1}} is not satisfied or \textbf{\textit{Condition 1}} and \textbf{\textit{Condition 2}} are satisfied simultaneously. Finally, the points are assigned to initial clusters with labels. The clusters that contain truly labeled points are assigned with pseudo labels which are the same as the label of the contained point. The simple algorithm is summarized in Algorithm \ref{alg_clustering}. The loop will terminate if any one of the convergence conditions is satisfied, which can be summarized as:
 \begin{enumerate}
     \item All points have been assigned with labels;
     \item There are no seed points that can be added;
     \item Regions will not expand between two successive steps.
 \end{enumerate}

As shown in Fig. \ref{Clust}, after region expansion based clustering, although the plane can be precisely segmented, points that are of the same semantic/instance are also partitioned, which means only employing geometry information will result in points being over-clustered. Therefore, the learnt semantics should be taken into account when generating pseudo labels. We propose to learn to merge over-divided clusters based on the local geometric similarity and the learnt feature similarity. As shown in the third column of Fig. \ref{Clust}, it turns out the generated pseudo labels by our proposed \textit{Cluster-Level Similarity Prediction Network} are of high quality thanks to iterations of optimization with our elaborately designed network modules, which is illustrated in Subsection B. Thanks to our fast implementations of Octree based KNN search, We can do region expansion and merging in real-time with training of networks iteratively, which significantly improve the speed of our framework.

\begin{figure*}[htb]
\setlength{\abovecaptionskip}{-0.15cm}
\setlength{\belowcaptionskip}{-0.15cm}
\centering
\includegraphics[scale=0.23]{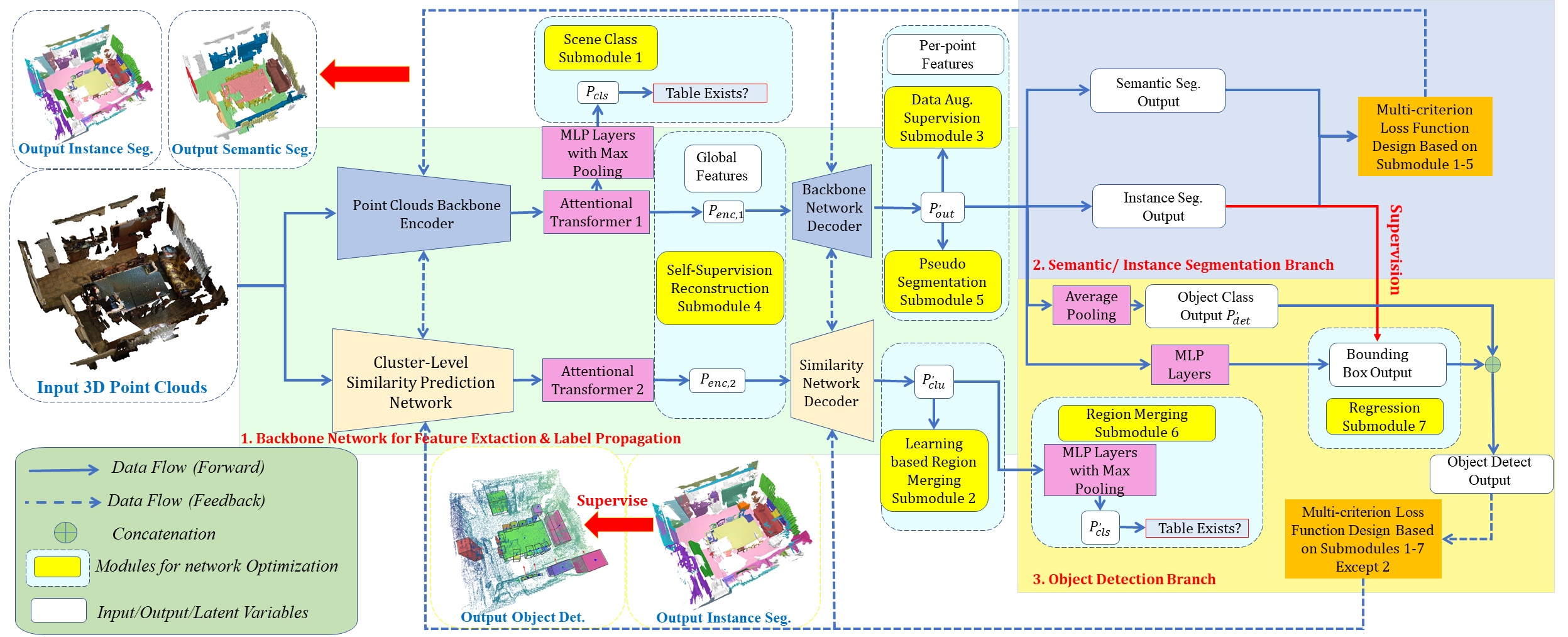}
\caption{\textbf{WeakLabel3D-Net Architecture overview}. The two backbone network adopts the same encoder-decoder structure to obtain the global and per-point features. The semantic/instance segmentation branch and object detection branch are the two main output branches supervised by our proposed network optimization modules. Our framework can be integrated seamlessly to any off-the-shelf point or voxel based backbones.}
\label{fig_backbone}
\vspace{-0.46cm}
\end{figure*}


\subsection{The Overall Network Architecture}
As shown in Fig. \ref{fig_backbone}, our network framework mainly consists of two network backbones, followed by two multi-task branches for 3D semantic/instance segmentation and object detection respectively. We propose submodules to optimize the framework in an end-to-end manner.
\subsubsection{Network Backbone}
 The network backbone is selected as our recently proposed FG-Net\cite{liufgnet}. Note that the backbone can be substituted by any off-the-shelf ones because our method is backbone-agnostic and can fit in seamlessly to any point-based or voxel-based network. In this work, We choose FG-Net for its strong capacity to modeling the local geometry and capturing both the geometric and semantic correlations. We adopt a cluster-level similarity prediction network in a complementary manner to FG-Net to obtain the similarities scores among clusters, which is detailed later in this Subsection. 
 
\subsubsection{Semantic/Instance Segmentation Branch}
The semantic/instance segmentation branch consists of the following submodules:

\textbf{Scene Class Submodule} The semantic/instance category existing within a scene can serve as global guidance to optimize the network. We directly utilize the global feature to predict whether an object presents within a scene or not, The loss function is formulated as:
\begin{equation}
        L_{Cls}=-\frac{1}{C_{cls}}\sum_{i=1}^{C_{cls}} \textit{\textbf{CE}}(\textbf{P}_{cls}, \textbf{P}^{\mbox{\textit{\tiny{GT}}}}_{cls})
\vspace{-1.6mm}
\end{equation}
Where $\textbf{P}_{cls}$, $\textbf{P}^{\mbox{\textit{\tiny}}{GT}}_{cls}$ are the prediction and ground truth for the presence of objects within a scene,  $C_{cls}$ is the number of semantic/instance classes with a scene. \textit{\textbf{CE}} stands for cross entropy loss.

\textbf{The Cluster-Level Similarity Prediction Network} The output of FG-Net gives the prediction of semantic segmentation with $\textbf{P}_{out} \in \mathbb{R}^{N_i \times C_{seg}}$, where $C_{seg}$ denotes the number of semantic/instance categories. For the limited annotation case, which means there are only a few annotated points (20-200 points) (i.e. 0.01\textperthousand\\-0.01\%) in a scene with approximately $2 \times 10^6$ points, we design a \textit{Cluster-Level Similarity Prediction Network} with different weights to serve as a contrastive module to offer clustering-level predictions, which has almost the identical architecture to FG-Net. However, We change $1 \times 1$ convolution at the last layer of FG-Net to obtain a feature $\textbf{P}^{'}_{clu} \in \mathbb{R}^{N_{clu} \times C_{Seg}}$, where $N_{clu}$ is the initial number of clusters obtained from region expanding. Note that normalized scores of merged clusters is added in each training iteration with merging of clusters based on the similarity scores in both geometry and semantics among them. We utilize $\textbf{P}^{'}_{clu}$ to give a category prediction at the cluster level instead of at the point level. To be more specific, each element $\textbf{p}_{clu, i} \in \mathbb{R}^{1\times C_{Seg}}$ indicates the class of the $i_{th}$ cluster. Utilizing the class of truly annotated points $\textbf{P}^{\mbox{\textit{\tiny{GT}}}}_{clu}$ as the ground truth class of the clusters containing them, the cross entropy loss is as follows to supervise cluster-level predictions:
\begin{equation}
    L_{Clus}=-\frac{1}{N_{label}}\sum_{i=1}^{N_{label}} \textit{\textbf{CE}}(\textbf{P}^{'}_{clu}, \textbf{P}^{\mbox{\textit{\tiny{GT}}}}_{clu})
\vspace{-1.6mm}
\end{equation}
Where $N_{label}$ is numbers of points in all clusters containing the truly annotated points, which will also change in each training iteration with merging of clusters. 

\textbf{Learning Based Region Merging Submodule}
The initial clusters obtained from region expansion suffer from over-segmenting or inaccurate partitioning. It is desired that a region merging submodule should be proposed to merge or divide clusters in a learnable way. In our design, the predicted semantic/instance of the learnable network and the geometric properties of clusters jointly decide a similarity score, indicating whether neighbouring clusters should be merged or partitioned. More specifically, the similarity score between two clusters $\textbf{P}_{clu, i}$ and $\textbf{P}_{clu,j}$ is calculated as: 
\begin{equation}
    S_{i, j}(\textbf{P}_\textit{\textbf{clu,i}}, \textbf{P}_\textit{\textbf{clu,j}})= y_1 M_{\mbox{\textit{\tiny{{color,i,j}}}}} +  y_2 M_{\mbox{\textit{\tiny{{scale,i,j}}}}}+ y_3 M_{\mbox{\textit{\tiny{{seg,i,j}}}}}+ y_4 M_{\mbox{\textit{\tiny{{iou,i,j}}}}}
\vspace{-1.6mm}
\end{equation}
Where $M_{\mbox{\textit{\tiny{{color,i,j}}}}}, M_{\mbox{\textit{\tiny{{scale,i,j}}}}}, M_{\mbox{\textit{\tiny{{iou,i,j}}}}}, M_{\mbox{\textit{\tiny{{seg,i,j}}}}} \in \{0, 1\}$. The $M_{\mbox{\textit{\tiny{{color,i,j}}}}}$, $M_{\mbox{\textit{\tiny{{scale,i,j}}}}}$, and $M_{\mbox{\textit{\tiny{{iou,i,j}}}}}$ are the scores that are the normalized average intrinsic color, dimension, and position similarities between two point clusters, respectively. While the semantic similarity $M_{\mbox{\textit{\tiny{{seg,i,j}}}}}$ between the $i_{th}$ and $j_{th}$ cluster should be learnt based on output similarity of the two clusters by \textit{Cluster-level Similarity Prediction Network}:
\begin{equation}
M_{seg}=\frac{1}{N_{\mbox{\textit{\tiny{clu,i,j}}}}}\sum_{i=1}^{N_{clu}}\sum_{i=j}^{N_{clu}-1}\frac{1}{\Vert\textbf{p}_{clu, i}- \textbf{p}_{clu, j}\Vert}
\vspace{-1.6mm}
\end{equation}
Where $N_{\mbox{\textit{\tiny{clu,i,j}}}}=N_{clu} (N_{clu}-1)$, $\textbf{p}_{clu, i}$ and $\textbf{p}_{clu, j}$ are the corresponding predictions in $\textbf{P}^{'}_{clu}$ for cluster $i$ and $j$. The balancing weights $y_1, y_2, y_3 \in \{0, 1\}$ are set to values declining from a high value to a small value, i.e. $y_1=y_2=y_3=1-\frac{m_i}{{{N_{Total}}}}$, while $y_4 \in \{0, 1\}$ is set to the value $\frac{m_i}{N_{Total}}$. Where $m_i$ represents the current iteration in training, and $N_{Total}$ is total number of training iterations. This design means we firstly trust similarity of geometric properties and gradually trust more on the updated semantics relations by Cluster-Level Prediction Network. We replace the \textbf{\textit{Condition 1}}  and \textbf{\textit{Condition 2}} in Algorithm \ref{alg_clustering} with
the \textbf{\textit{Condition 3}}: $S_{i, j} \geq 1.25$ and \textbf{\textit{Condition 4}}: $S_{i, j} \geq 1.5$ respectively. Also, we substitute K nearest neighbour (KNN) points in Algorithm \ref{alg_clustering} with KNN clusters. This module, combined with \textit{Cluster-level Similarity Prediction Network}, utilizes weak labels as the guidance to increase the quality of generated pseudo label in training iteratively for both semantic and instance segmentation tasks.


\textbf{Data Augmentation Supervision Submodule}
This submodule is inspired by a simple intuition that the network prediction should be consistent under diverse transformations including flipping, rotation, and even down-sampling. We firstly obtain the final network semantic/instance predictions $\textbf{P}^{'}_{out}, \;\textbf{P}^{aug}_{out} \in \mathbb{R}^{N_i \times C_{seg}}$ respectively. The \textit{KL} divergence is universally adopted to evaluate the difference between two Probabilistic distributions. In our work, we utilize the \textit{JS} divergence instead because of its symmetry property, which make it remain constant when two distributions are distant to each other. The final \textit{JS} Divergence Loss is formulated as:
\begin{equation}
\begin{aligned}
    L^{js}_{Aug} =& -\frac{1}{N_{com}}\sum_{i=1}^{N_{com}} \textbf{Div}_{\small{JS}} (\sigma(\textbf{P}^{'}_{out})\|\sigma(\textbf{P}^{aug}_{out}))
\vspace{-1.6mm}
\end{aligned}
\end{equation}
Where $N_{com}$ is the intersectional common points between $\textbf{P}^{'}_{out}$ and $\textbf{P}^{aug}_{out}$, and $\sigma$ is the softmax function with normalization to produce probabilistic scores for each class. After applying the data augmentation constraints, we aim at ensuring that the distribution of the probabilistic scores of segmentation will remain consistent between the  $\textbf{P}^{'}_{out}$ and  $\textbf{P}^{aug}_{out}$ after various data transformations. To be more specific, transformation invariance can be achieved.

\textbf{Self-Supervision Reconstruction Submodule}
It can be observed that the scene contains many isolated point cloud clusters which belongs to the same semantic category. The limited labels in the current labeled clusters can be propagate to semantically similar unlabled clusters to provide supervisions. We first use the transformer module to find the semantically similar regions. Then we reconstruct the semantic prediction from the redistributed feature. It is apparent that the correctly redistributed feature should be capable of offering the correct semantic/instance segmentation as well. Therefore, denote the encoded feature in the latent space as  $\textbf{P}_{enc} \in \mathbb{R}^{N_{enc} \times C_{enc}}$, we first apply a transformer layer with weight $\textbf{W}_\textbf{{trans1}}, \textbf{W}_\textbf{{trans2}} \in \mathbb{R}^{C_{enc} \times C_{mid}}$  to transform $P_{enc}$ into latent representations $\textbf{P}_{mid1},\textbf{P}_{mid2} \in \mathbb{R}^{N_{enc} \times C_{mid}}$.  Then $\textbf{P}_{mid1}, \textbf{P}_{mid2} $ can be multiplied to obtain a transformer matrix $\textbf{W}_\textbf{{A}}=\textbf{P}_{mid1}\textbf{P}_{mid2} ^{\mbox{\textit{\tiny{T}}}}\in \mathbb{R}^{N_{enc} \times N_{enc}}$,  which can be regarded as the weight to be learnt by network. Then $\textbf{W}_\textbf{{A}}$ is normalized by softmax into a score matrix $\textbf{W}^{'}_\textbf{{A}} \in \mathbb{R}^{N_{enc} \times N_{enc}}$. We multiply $\textbf{W}^{'}_\textbf{{A}}$ with the original $\textbf{P}_\textbf{enc}$ to obtain a new latent representation $ \textbf{P}_\textbf{enc,1} \in \mathbb{R}^{N_{enc} \times C_{enc}}$.  Note that our operation will not add extra computation cost because transformer operations has already been realized in FG-Net\cite{liufgnet}. Different from FG-Net, we reconstruct two semantic prediction directly by $\textbf{P}_\textbf{enc}, \textbf{P}^{'}_\textbf{enc} $, and obtain two network output $\textbf{P}_\textbf{out} ,\textbf{P}^{'}_\textbf{out}  \in \mathbb{R}^{N_{i} \times C_{seg}}$.  We use the \textit{JS} divergence to evaluate the difference between the two distributions, and make it smaller with reconstruction loss in optimization:
\begin{equation}
\begin{aligned}
    L^{js}_{Rec} =& -\frac{1}{N_{i}}\sum_{i=1}^{N_{i}} \textbf{Div}_{\small{JS}} (\varphi(\textbf{P}_{enc})\|\varphi(\textbf{P}^{'}_{enc}))=  \\ & -\frac{1}{N_{i}}\sum_{i=1}^{N_{i}} \textbf{Div}_{\small{JS}} (\textbf{P}_{out}\|\textbf{P}^{'}_{out})
\end{aligned}
\vspace{-2.2mm}
\end{equation}
Where $\varphi$ stands for the decoder of FG-Net. As transformers can be regarded as a transformation in the embedding space, The cross entropy loss is also applied on $\textbf{P}_{out}$ to propagate weak labels to the semantically similar points, given as:
\begin{equation}
    L_{Att}=-\frac{1}{N_{i}}\sum_{i=1}^{N_{i}} \textit{\textbf{CE}}(\textbf{P}_{out}, \textbf{P}^{\mbox{\textit{\tiny{GT}}}}) \mathds{1}(\textbf{p}_{i})
\vspace{-1.6mm}
\end{equation}
Where $\mathds{1}(\textbf{p}_{i}) \in \{0, 1\}$ indicates whether the point is a truly labeled point. In this way, we encourage similar features in the latent space to have a similar probabilistic distribution over semantic/instance predictions, and the instance labels can successfully spread to the similar clusters whose features are related in the encoding space.


\begin{figure*}[htbp]
\setlength{\abovecaptionskip}{-0.15cm}
\setlength{\belowcaptionskip}{-0.15cm}
\centering
\includegraphics[scale=0.479]{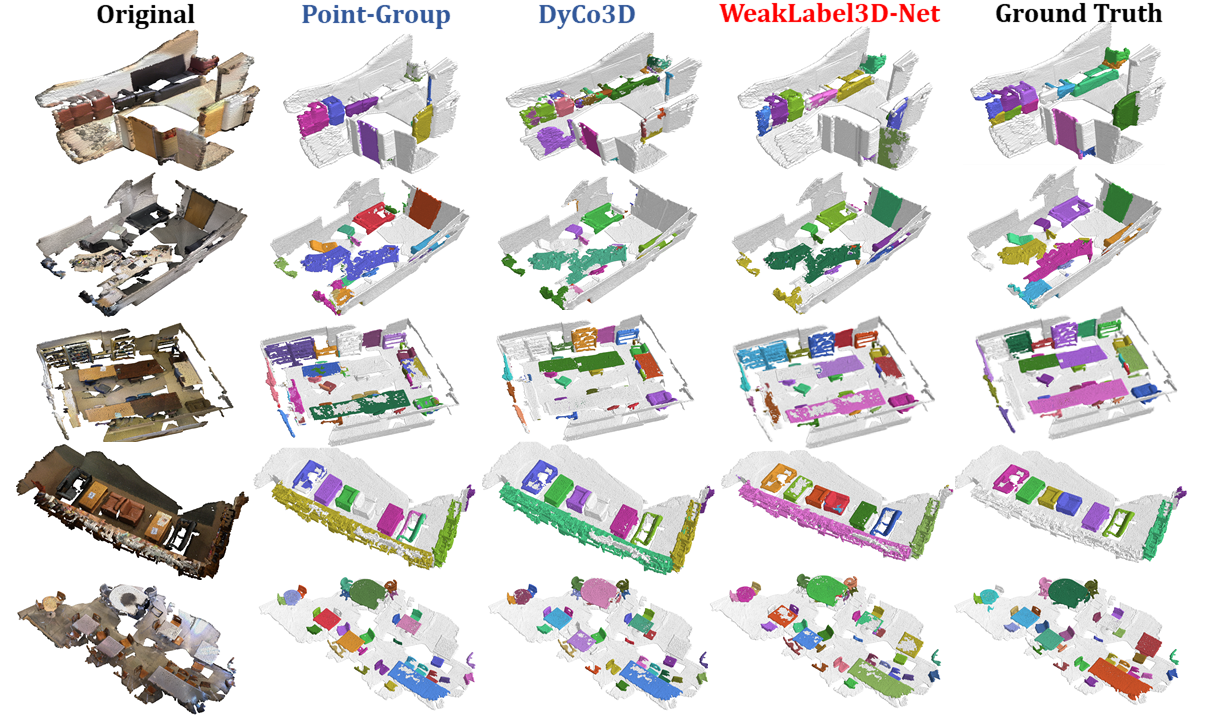}
\caption{Qualitative \textbf{instance segmentation} results on ScanNet compared with fully supervised methods with different instances indicated by different colors. The wall and floor are set to white because they are of the same instance, which is not counted for comparisons in public online benchmark.}
\label{fig_scannet}
\vspace{-0.39cm}
\end{figure*}

\textbf{Pseudo Segmentation Submodule}
Finally, the network is also guided by the generated pseudo label in both semantic and instance segmentation tasks, which can be formulated as:
$ L_{\mbox{\textit{\tiny{{WSL}}}}}=\frac{1}{N_{i}}\sum_{i=1}^{N_{i}}\textit{\textbf{CE}}(\textbf{P}^{'}_{out}, \textbf{P}_{gt}) \mathds{1}(\textbf{p}_{i})$. Where $\textbf{P}^{'}_{out}$ is the segmentation output prediction, and $\textbf{P}_{gt}$ is the ground truth supervision provided by the generated pseudo labels in each training iterations. $\mathds{1}(\textbf{p}_{i}) \in \{0, 1\}$ indicates whether the point has been given a pseudo label. The final optimization takes losses from all above-mentioned submodules into account, formulated as $ L_{Seg}=L_{Cls}+L_{Clus}+L^{js}_{Aug}+L^{js}_{Rec}+L_{Att}+L_{\mbox{\textit{\tiny{WSL}}}}$. The network is optimized in an end-to-end manner for semantic/instance segmentation tasks.
\subsection{Object Detection Branch}
\textbf{Region Merging Submodule Specially for Detection}
Object detection can take advantage of the supervision from instance segmentation because object proposals can be directly obtained from instances. The axis tightly aligned bounding box of each instance is selected as the initialization of pseudo ground truth bounding boxes. Detection network is designed on widely adopted votenet, but \textbf{Dice} loss is designed to guarantee tighter aggregations of points within the same cluster, and strict geometric separations of points in diverse clusters. Note that for object detection, our method operates in a unsupervised manner for instance segmentation, followed by our regression submodule to realize object detection. Different from segmentation branch, $\textbf{P}^{'}_{out} \in N_i \times (C_{det}+1)$, where $(C_{det}+1)$ is the number of classes plus one for backgrounds in detection. We apply average pooling to $\textbf{P}^{'}_{out}$ to obtain $\textbf{P}^{'}_{det} \in N_{\mbox{\textit{\tiny{{R}}}}} \times (C_{det}+1)$, where $N_{\mbox{\textit{\tiny{{R}}}}}$ is the number of pseudo labeled clusters, which is corresponding to pseudo labeled bounding boxes. In the region merging submodule, we add $1\times 1$ convolution and max pooling after $\textbf{P}^{'}_{clu}$ of the similarity prediction network to produce $\textbf{P}^{'}_{cls} \in \mathbb{R}^{N_{cls}} $ to predict the presence of object or not with a scene: 
    \begin{equation}
        L_{cls,2}=-\frac{1}{C_{cls}}\sum_{i=1}^{C_{cls}} \textit{\textbf{CE}}(\textbf{P}^{'}_{cls}, \textbf{P}^{GT}_{cls})
 \vspace{-1.6mm}
 \end{equation}
    In this way, object presence within a scene can serve as the supervision for similarity predictions among clusters benefiting from self-supervision by scene object classes. And proposals can be merged with merging of point instances in instance segmentation.

\begin{table*}[tb]
\setlength{\abovecaptionskip}{-0cm}
\setlength{\belowcaptionskip}{-0cm}
\tiny
\caption{The Comparisons of the performance of our proposed method on various of benchmarks. For Weakly Supervised Semantic/Instance Segmentation, the test circumstance of \textbf{20 labeled points} is showed on ScanNet for comparisons}
\label{tablewslresults}
\begin{center}
\begin{tabular}{ccccccccccccccccc}
\toprule
\multirow{2}{*}{Method} & \multicolumn{3}{c}{ScanNet Semantic Seg.\%} & \multicolumn{3}{c}{ScanNet Instance Seg.\%} & \multicolumn{3}{c}{ScanNet Object Det.\%}& \multicolumn{2}{c}{S3DIS Semantic Seg.\%}& \multicolumn{2}{c}{S3DIS Instance Seg.\%}& \multicolumn{3}{c}{KITTI Object Det.(Car)\% .} \cr &Average IOU & Bathtub&bed & AP50 & bed &bookshelf & AP50 & bed &bookshelf &mIOU& wall& mPrec &mRec& moderate & easy& hard
\\

\hline
WeakLabel-3DNet (Ours)&\textcolor{red}{\textbf{66.1}}&80.8&77.1&\textcolor{red}{\textbf{55.1}}&69.2&48.1

&\textcolor{red}{\textbf{35.9}}&63.5&43.3&\textcolor{red}{\textbf{66.5}}&79.5&\textcolor{red}{\textbf{65.2}}&\textcolor{red}{\textbf{48.2}}&91.2&95.7&86.3\\ 

One-Thing-One-Click\cite{liu2021one}&58.7&76.3&73.1&47.3&58.2&42.1

&25.2&51.4&36.0&55.9&63.6&57.5&41.3&80.5&85.1&76.2\\

PointContrast\cite{xie2020pointcontrast}&55.3&72.8&68.1&29.5&29.4&34.3

&15.3&29.8&16.8&30.3&43.3&51.8&37.9&72.1&76.9&67.5\\

Viewpoint-Bottleneck	\cite{sun20203d}&54.5&75.2&58.3&36.5&46.7&34.5

&22.3&36.9&23.5&38.3&45.5&49.8&33.5&69.9&75.2&63.1\\

ContrastiveSceneContext\cite{hou2021exploring}&54.2&65.7&63.5&29.2&58.1&41.9

&18.9&35.6&12.1&39.9& 50.1&52.9&43.8&76.4&82.3&71.9\\

Scratch\_CSC \cite{hou2021exploring}&37.9&38.6&60.7&20.1&63.6&18.4

&9.9&25.5&8.5&28.4&28.2&23.6&15.4&50.3&55.0&43.5\\
\bottomrule
\end{tabular}
\end{center}
\vspace{-0.6cm}
\end{table*}



    





 \textbf{Regression Submodule}
Note that similar from segmentation branch, the self-supervision submodule and data augmentation submodule are utilized for predict the class of the bounding boxes. 
At the same time, the \textbf{Dice Loss} can be utilized to evaluate intersections between the predicted clusters and ground truth clusters for regression purpose. Note that other submodules are the same as the semantic/instance segmentation branch. Denote the \textbf{Dice loss} as $L_{Dice}$ and the same losses as segmentation branch as $L_{Seg,2}$, the optimization function for detection is formulated as $L_{Det}=L_{Seg,2}+L_{Dice}$. Network is optimized in an end-to-end manner on a single 1080Ti GPU for three scene understanding tasks.
\section{Experiments}
\subsection{Experimental Details}
 The network is trained for 300 epoches on a single 1080Ti with batch size of 8 during training and 16 during testing. The initial learning rate is $1e^{-3}$ and decays by 5 times every 60 epoches. We implemented it in \textit{PyTorch} and optimized it with Adam optimizer. 
 \vspace{-0.1cm}
\subsection{Results of WSL for 3D Semantic/Instance Segmentation}
\begin{figure}[htbp]
\setlength{\abovecaptionskip}{-0.12cm}
\setlength{\belowcaptionskip}{-0.3cm}
\centering
\includegraphics[scale=0.20]{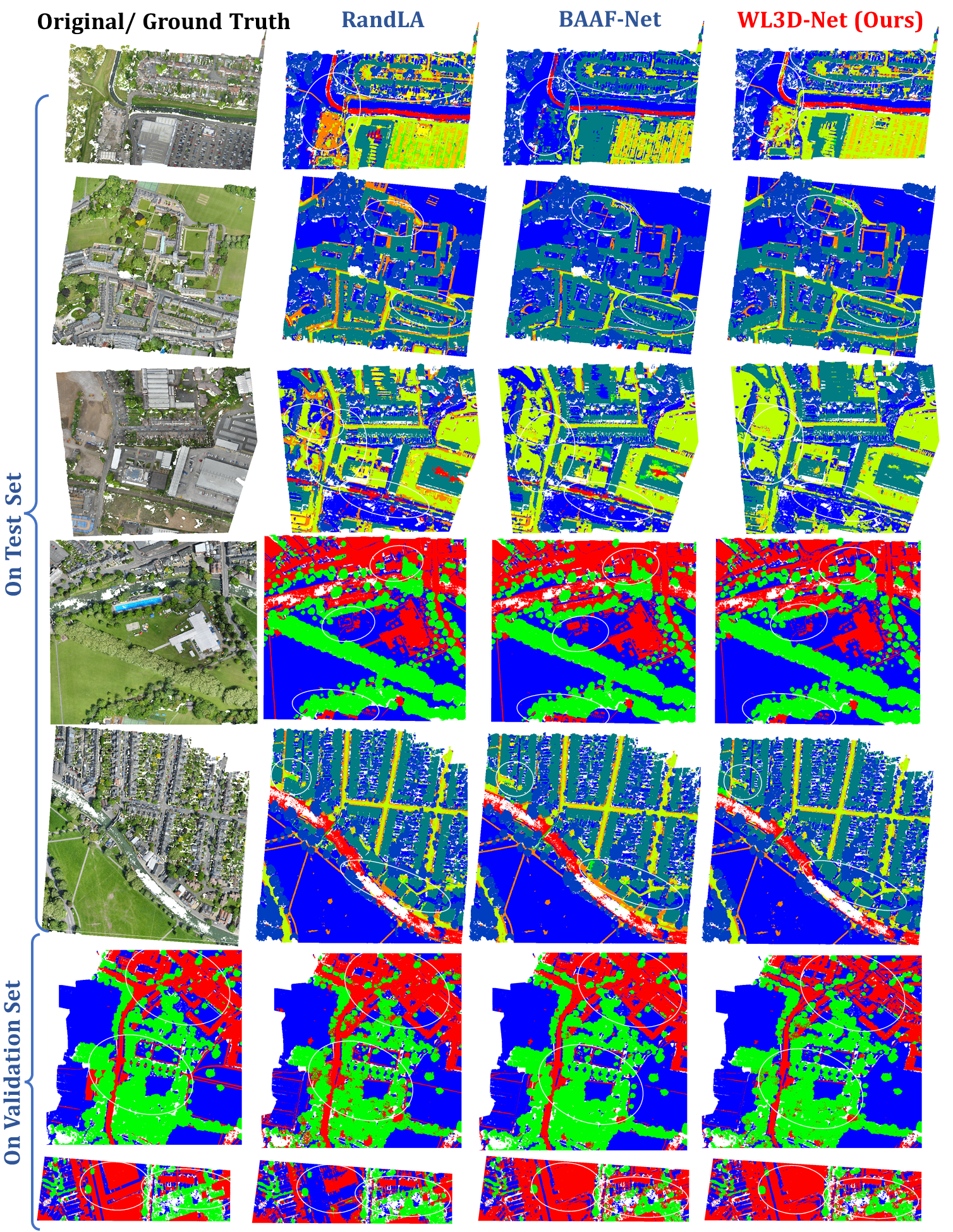}
\caption{Results of large-scale \textbf{semantic segmentation} performance compared with fully supervised methods RandLA and BAAF-Net. Different Color indicates different Semantics. The first 5 rows show results on test set and the last 2 rows show results on validation set. The ground truth is only available for val. set. Red stands for our WSL method while blue stands for fully supervised methods. White circles highlight differences in predictions.}
\label{fig_sensat}
\vspace{-3.9mm}
\end{figure}

Our framework is tested extensively on various large-scale point clouds understanding benchmarks including S3DIS\cite{armeni20163d}, SensatUrban \cite{hu2021towards} and ScanNet for 3D Semantic/Instances Segmentation under supervisions with \{20, 50, 100, 200\} points, which means the label percentage range from 0.01\textperthousand\ to 0.01\%. As shown in Table \ref{tablewslresults}, our framework \textbf{ranks first} in the task of semantic/instance segmentation with limited annotations. Remarkable performance has been achieved in most semantic/instance categories, which outperforms current art One Thing One Click\cite{liu2021one} by 6.8\% for semantic segmentation and outperforms Contrastive Scene Context by more than 20\% for instance segmentation on online public benchmark. We have also tested our method at the current biggest urban-level large-scale point clouds segmentation benchmark SenSat-Urban, the results are shown in Fig. \ref{fig_sensat}. It can be seen that our method can offer comparable results with those with full supervisions, such as RandLA and current SOTA BAAF-Net. we achieve mIOU of 56.8\% on validation sets, which is comparable with performance of fully supervised counterparts RandLA (55.9\%) and BAAF-Net (58.3\%). 
%
\begin{figure}[htbp]
\setlength{\abovecaptionskip}{-0.12cm}
\setlength{\belowcaptionskip}{-0.3cm}
\centering
\includegraphics[scale=0.3]{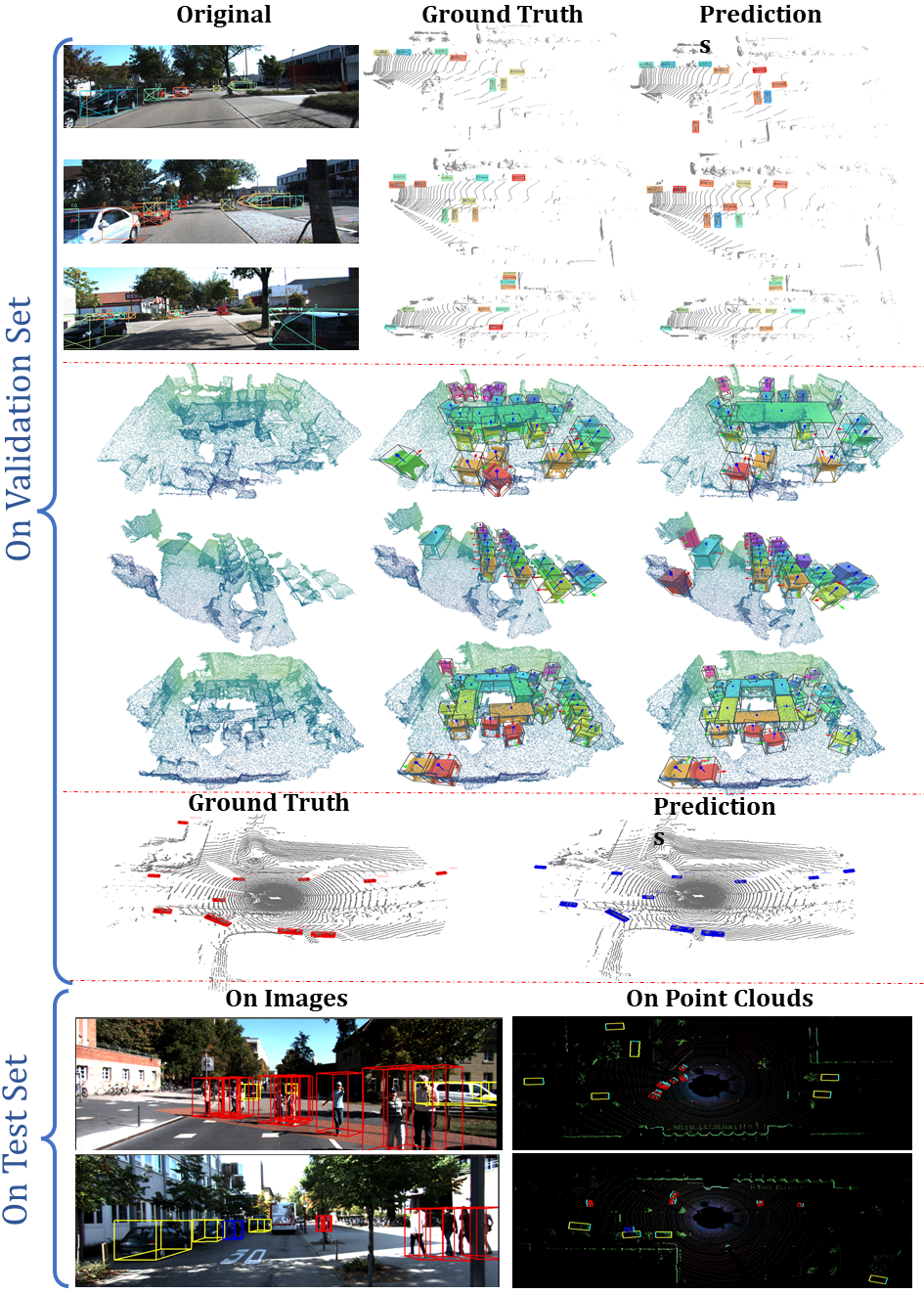}
\caption{\textbf{Object detection} results on test and validation set for KITTI/ScanNet/Waymo benchmarks respectively. The row 1-3, and 8-9 are results on KITTI; Row 4-6 and row 7 are results for ScanNet and Waymo respectively.}
\label{fig_all}
\vspace{-5.99mm}
\end{figure}
 For instance segmentation, the qualitative results of Scannet Instance Segmentation is shown in Fig. \ref{fig_scannet} in contrast to those fully-supervised baselines DyCo3D\cite{he2021dyco3d} and Point-Group\cite{jiang2020pointgroup}, and quantitative comparisons are given in Table \ref{tablewslresults}. We have tested our methods in diverse circumstances with 20 labeled points to 200 labeled points respectively with the metrics of AP (Average Precision), AP 50\%, and AP 25\%. It turns out our method ranks first in the instance segmentation with \{20, 50, 100, 200\} labeled points simutaneously, improved on current weakly supervised SOTA ContrastiveSceneContext\cite{hou2021exploring} by a great margin of at least 10\%, which demonstrates the superior performance of our framework. We have also tested our method in a complete unsupervised manner only with scene object class level labels. We still reach AP of 46.8\% in the AP 50\% scenario for ScanNet, which demonstrates our hypothesis that the our instance segmentation results are accurate enough to instruct and provide supervision for the object detection.
\subsection{Results of WSL for KITTI/Waymo/ScanNet Object Detections}
We have also done experiments of object detection on ScanNet, KITTI\cite{geiger2013vision}, and Waymo\cite{sun2020scalability} benchmarks. It can be demonstrated that our framework outperforms other weakly supervised counterparts by a large margin. The qualitative visualizations of validation set outcomes on three large-scale benchmarks is shown in Fig. \ref{fig_all} and quantitative results are summarized in Table \ref{tablewslresults}. It is apparent that our framework provides consistent excellent performance for 3D semantic/instance segmentation and 3D object detection simultaneously, and its effectiveness for both indoor and outdoor 3D scene understanding is further demonstrated. 
 \vspace{-0.23cm}

\section{Conclusion}
In conclusion, we have proposed a general benchmark framework for weakly supervised point clouds understanding which has superior performance for the three most significant semantic understanding tasks including \textit{3D Semantic/Instance Segmentation} and \textit{Object Detection}. The proposed network learns to merge over-divided clusters based on the local geometric property similarities and the learnt feature similarities. Network modules are proposed to fully investigate the relations among semantics within a scene, thus, high-quality pseudo labels can be generated from weak labels to provide a better segmentation supervision. The effectiveness of our approach is verified across diverse large-scale real-scene point clouds understanding benchmarks under various test circumstances. Our label-efficient learning framework has great potentials for robotic 3D scene understanding tasks when labels are inaccessible or difficult to be obtained, such as robot explorations in complex scenarios, or robot indoor/outdoor interactions with environment.

\addtolength{\textheight}{0cm}   





\bibliographystyle{IEEEtran}
\bibliography{references}

\end{document}